\relax
\documentclass[letterpaper]{article}
\usepackage{arxiv}
\usepackage{times}
\usepackage{helvet}
\usepackage{courier}
\usepackage[hidelinks]{hyperref}       
\usepackage{graphicx}
\usepackage{graphicx}
\newcommand\blfootnote[1]{%
  \begingroup
  \renewcommand\thefootnote{}\footnote{#1}%
  \addtocounter{footnote}{-1}%
  \endgroup
}

\usepackage{booktabs}       
\usepackage{amsfonts}       
\usepackage{nicefrac}       
\usepackage{amsmath}
\usepackage{enumitem}

\begin{document}
\title{The Missing Margin: How Sample Corruption Affects Distance to the Boundary in ANNs}

\author{
  Marthinus W. Theunissen$^{1, 2, *}$, Coenraad Mouton$^{1, 2, 3, *}$, Marelie H. Davel$^{1, 2, 4}$\\ \\
    $^{1}$Faculty of Engineering, North-West University, South Africa\\
    $^{2}$Centre for Artificial Intelligence Research (CAIR), South Africa\\ $^{3}$South African National Space Agency (SANSA), South Africa\\ $^{4}$National Institute for Theoretical and Computational Sciences (NITheCS), South Africa\\
  \{tiantheunissen, moutoncoenraad, marelie.davel\}@gmail.com \\
}

\maketitle

\blfootnote{This work is a preprint of a published paper by the same name~\cite{theunissen2022missing}, which it subsumes. This preprint is an {\it extended version}: it contains additional empirical evidence (clean and input-corrupted CNN models) and additional discussion. The reduced, authenticated version is available online at \url{https://doi.org/10.1007/978-3-031-22321-1_6}.}

\def\thefootnote{*}\footnotetext{Equal contribution.}

\begin{abstract}
Classification margins are commonly used to estimate the generalization ability of machine learning models. We present an empirical study of these margins in artificial neural networks. A global estimate of margin size is usually used in the literature. In this work, we point out seldom considered nuances regarding classification margins. Notably, we demonstrate that some types of training samples are modelled with consistently small margins while affecting generalization in different ways. By showing a link with the minimum distance to a different-target sample and the remoteness of samples from one another, we provide a plausible explanation for this observation. We support our findings with an analysis of fully-connected networks trained on noise-corrupted MNIST data, as well as convolutional networks trained on noise-corrupted CIFAR10 data.
\end{abstract}


\section{Introduction}
\label{introduction}


The study of artificial neural networks (ANNs) and their performance on unseen test data embodies many different overlapping themes, such as capacity control \cite{vapnik1999overview}, loss landscape geometry \cite{hochreiter1997flat}, information theoretical approaches \cite{tishby2015deep}, and algorithmic stability \cite{bousquet2000algorithmic}. Optimal \textit{classification margins} remains a popular concept, showing both theoretical support and empirical evidence of being related to generalization \cite{predict_gen_margin,rep_based_complexity_pgdl}. Simply put, the classification margin of a sample with regard to a specific model is the shortest distance the sample will need to move, in a given feature space, in order to change the predicted output value. Hereafter, we refer to this concept as a \textit{margin}. In the literature it is also called `minimum distance to the boundary' or `minimum adversarial perturbation'~\cite{moosavi2016deepfool}, depending on context.

Large margins have long been used to indicate good generalization ability in classification models~\cite{boser1992training,weinberger2009distance}. The supporting intuition is simple: With a larger margin, a sample can have more varied feature values (potentially due to noise) while still being correctly classified. 
It is argued that overparameterized ANNs tend to find large margin solutions~\cite{lyu2019gradient,wei2018margin}. 

Recent studies of the relationship between margin and generalization typically measure the average margins over the training set or some sampling thereof. 
In this work, we ask whether the margins of individual samples tend to reflect this average behaviour.
Specifically, we focus on ANN-based classification models and 
introduce controlled noise into the training process. 
Using different types of training sample corruption, we demonstrate a number of intricacies related to input margins and their relation to generalization. 
Specifically, we contribute the following:
\begin{enumerate}
    \item By using target and input noise, we point out local margin behaviour that is inconsistent with the global average. We find that, while all margins have a strong tendency to increase, label- and input-corrupted samples maintain significantly smaller margins than uncorrupted samples. We also find that only on-manifold corrupted samples (i.e. label corruption) noticeably affect the margins of clean samples.
    \item We discuss the implications of these inconsistencies. Our findings suggest that using the average margins as a metric is only fitting if the set of contributing samples has an equal level of diversity for all models being compared.  
    \item We probe the underlying mechanisms that lead to these inconsistencies. We hypothesize that label-corrupted samples have reduced margins because of their proximity to clean samples, and the input-corrupted samples have smaller margins due to a lack of incentive to increase them. 
\end{enumerate}
Our choice of using noise is not arbitrary. Adding artificial noise to a training set and investigating the ability of ANNs to generalize in spite of this corruption is a popular technique in empirical investigations of generalization. A good example of success with such methods is the seminal paper by Zhang et al.~\cite{zhang2021understanding}, where it was shown that overparameterized models can generalize in spite of having enough capacity to fit per-sample random noise such as label corruption or randomized input features. Similar noise has been used extensively to experimentally probe ANNs~\cite{neyshabur2017exploring,nakkiran_dd,theunissen2020benign}.

The rest of the paper is structured as follows: Section \ref{related work} describes related work on classification margins and generalization. In Section \ref{sec:measuring_classification_margins} we define a margin and describe our exact method of measuring it. Following this, Section \ref{experimental setup} presents details on our experimental setup. The resulting margins, along with notable local inconsistencies, are presented and discussed in Section \ref{local inconsistencies}. In the final section, we investigate the nature of these local inconsistencies, describing plausible underlying mechanisms.

\section{Related Work}
\label{related work}
Research on margins extends back much earlier than the advent of powerful ANNs. For example, the effect of margins on generalization performance in linear models such as Support Vector Machines (SVMs) are well-studied~\cite{vapnik1999overview}. An inherent issue with extending this work to modern ANNs is that their decision boundaries are highly non-linear, high dimensional, and contain high curvature. 

Finding the closest boundary point to a given sample is often considered intractable, as such, some works opt to rather estimate the margin~\cite{sokolic2017robust,predict_gen_margin} or define bounds on these margins~\cite{sun2015large}.  Elsayed et al.~\cite{large_margin_dnns} derive a linear approximation of the shortest distance to the decision boundary.  This is then formulated as a penalty term which is used during training to ensure that each sample has at least a specific (chosen hyperparameter) distance to the decision boundary, for both the input space and hidden layers.  Networks trained using this loss function exhibit greater adversarial sample robustness, and better generalization when trained on data with noisy labels than networks trained using conventional loss functions.  Similarly, Jiang et al.~\cite{predict_gen_margin} further utilize the same approximation to predict the generalization of a large number of Convolutional Neural Networks (CNNs), by training a linear regression models on the margin distributions of the existing models. However, it is shown in \cite{constrained_optim} that this approximation likely considerably underestimates the true distance to the decision boundary.

Several authors~\cite{deep_dig,constrained_optim,guan_linear_interpol} use a simple linear interpolation between two samples of different classes to find a point on the decision boundary which separates them.  This is however unlikely to result in a distance that is near the true minimum.  In a similar approximate fashion, Somepalli et al.~\cite{twice_dd_real_ver} take the average of the distance in five random directions around an input sample, where each directional-distance is calculated separately using a simple bisection method.

Karimi et al.~\cite{deep_dig} introduce `DeepDIG', an approach based on an auto-encoder that finds samples near the decision boundary that are visually similar to the original training samples. However, this method does not specifically attempt to find the nearest point on the decision boundary.  Finally, Youzefsadeh and O'Leary~\cite{constrained_optim} formulate finding the shortest distance to the decision boundary of a given sample as a constrained optimization problem.  While this method is highly accurate, it is computationally very expensive, especially for high-dimensional data, e.g. natural images such as MNIST~\cite{Lecun98gradient-basedlearning} and CIFAR10~\cite{cifar}.

Similar to Youzefsadeh and O'Leary~\cite{constrained_optim}, in this work we find actual points in feature space. We have two reasons for this: (a) simplified approximations might lead to misconceptions of the role margins play in a model's ability to generalize, and (b) finding actual points inherently considers the non-linear intricacies in the function mapping. In Section \ref{sec:measuring_classification_margins} we describe our selected method in detail.

\section{Formulating the Classification Margin}
\label{sec:measuring_classification_margins}

In the previous section, we have given an overview of different methods of calculating the classification margin. We opt to use the most precise method, which formulates the margin calculation as a non-linear constrained minimization problem.

Let $f : X \rightarrow \mathbb{R}^{|N|}$ denote a classification model with a set $N = \{0, 1, ..., n\}$ of output classes.  For an input sample $\mathbf{x}$ we search for a point $\mathbf{\hat{x}}$ on the decision boundary between classes $i$ and $j$, where $i$ is $\arg \max (f(\mathbf{x}))$, with $i, j \in N, j \neq i$, and $\mathbf{\hat{x}}$ is the nearest point to $\mathbf{x}$ on the decision boundary. Formally, for some distance function $dist$, we find $\mathbf{\hat{x}}$ by solving the following constrained minimization problem (CMP):
\begin{equation}
\label{eq:objective_func}
    \min_{\mathbf{\hat{x}}} dist(\mathbf{x}, \mathbf{\hat{x}})
\end{equation}
such that
\begin{equation}
\label{eq:eq_constraint}
    f(\mathbf{\hat{x}})[i] = f(\mathbf{\hat{x}})[j] 
\end{equation}
for the $i^{th}$ and $j^{th}$ output node, respectively. Finding a point that meets the condition defined in Eq.~\ref{eq:eq_constraint} exactly, is virtually impossible. In practice, a threshold is used, so that a point is considered valid (on the decision boundary) if $|f(\mathbf{\hat{x}})[i] - f(\mathbf{\hat{x}})[j]| \leq 10^{-3}$.  In order to find the nearest point on the decision boundary for all $j$, we search over each class $j \neq i$ separately for each sample and choose the one with the smallest distance. As is convention for margin measurements \cite{constrained_optim,twice_dd_real_ver}, we use Euclidean distance\footnote{In practice, we optimize for the squared Euclidean distance in order to reduce the computational cost of gradient calculations, but report on the unsquared distance in all cases.} as metric, meaning the margin is given by:
\begin{equation}
dist(\mathbf{x}, \mathbf{\hat{x}}) =  |\mathbf{x} - \mathbf{\hat{x}}|_{2}
\end{equation}

In order to solve each CMP, we make use of the augmented Lagrangian method~\cite{nlopt_lagrange_2}, using the conservative convex separable approximation quadratic (CCSAQ) optimizer~\cite{ccsaq} for each unconstrained optimization step, implemented with the NLOpt (non-linear optimization) library~\cite{nlopt} in Python.

While it is possible to extend the nearest boundary optimization to the hidden layers, it is difficult to compare models with layers of varying dimensionality.  Additionally, hidden layer margins can be manipulated and require some form of normalization~\cite{predict_gen_margin}. As such, we limit our analysis to input-space margins.

\section{Experimental Setup}
\label{experimental setup}

In order to investigate how sample corruption affects margin measurements, we train several networks of increasing capacity to the point of interpolation (close to zero train error) on the widely used classification datasets MNIST~\cite{Lecun98gradient-basedlearning} and CIFAR10~\cite{cifar}. `Toy problems' such as these are used extensively to probe generalization~\cite{nakkiran_dd,twice_dd_real_ver,zhang2021understanding}.  We corrupt the training data of some models using two types of noise, defined in Section \ref{controlled noise}, separately. Capacity and generalization are strongly linked. In the overparameterized regime we expect generalization to improve systematically with an increase in capacity and a corresponding increase in average margins. This setup allows us to determine whether expected behaviour is consistent across all samples.

\subsection{Controlled Noise}
\label{controlled noise}

We use two specific types of noise, inspired by Zhang et al.~\cite{zhang2021understanding}: Label corruption and Gaussian input corruption. These have been designed to represent two complications that are often found in real world data and could affect the generalization of a model fitted to them. Label corruption represents noise that comes from mislabeled training data (mislabeling is common in large real-world datasets and even in benchmark datasets~\cite{northcutt2021pervasive}), inter-class overlap, and general low separability of the underlying class manifolds. 

Gaussian input corruption, on the other hand, represents extreme examples of out-of-distribution samples. These are `off-manifold' samples displaying a high level of randomness. Such samples do not necessarily obscure the true underlying data distribution, but still require a significant amount of capacity to fit the excessive complexity that needs to be approximated when fitting samples with few common patterns.

Given a training sample $(\mathbf{x}, c)$ where $\mathbf{x} \in \mathbb{R}^{d}$ and $c \in N$ for a set of classes $N$, the corruption of a sample can be defined as follows:
\begin{itemize}
    \item \textit{Label corruption}: $(\mathbf{x}, c) \rightarrow (\mathbf{x}, \hat{c})$ where $\hat{c} \neq c, \hat{c} \in N$.
    \item \textit{Gaussian input corruption}: $(\mathbf{x}, c) \rightarrow (\mathbf{g}, c)$ where $\mathbf{g} \in \mathbb{R}^{d}$ and each value in $\mathbf{g}$ is sampled from $\mathbb{N}(\mu_{\mathbf{x}}, \sigma_{\mathbf{x}})$.
\end{itemize}
Alternative labels are selected at random and $\mathbb{N}(\mu_{\mathbf{x}}, \sigma_{\mathbf{x}})$ is a normal distribution, with  $\mu_{\mathbf{x}}$ and $\sigma_{\mathbf{x}}$ the mean and standard deviation of all the features in the original sample $\mathbf{x}$. Henceforth, we will drop the `Gaussian' when referring to `Gaussian input corruption'.

\subsection{MNIST Models}

For the MNIST dataset, we train three distinct sets of Multilayer Perceptrons (MLPs), with each set containing models of identical depth and identically varied width: 
\begin{itemize}
    \item \textbf{MNIST}: A set of clean MNIST models. These serve as baselines, showing the level of generalization and margin sizes to be expected should the models not have been trained on any corrupted data. 
    \item  \textbf{MNISTlc} (MNIST-label-corrupted): Models with the same capacities as the previous set, but where $20\%$ of the training set is label corrupted.
    \item \textbf{MNISTgic} (MNIST-Gaussian-input-corrupted): Models with the same capacities as the clean models, however, $20\%$ of the training set is input corrupted.
\end{itemize}
All models for these tasks have the following hyperparameters in common. They use a $55\ 000/5\ 000$ train-validation split of the training data. They are all single hidden layer ReLU-activated MLPs with widths ranging from $100$ to $10\ 000$ hidden layer nodes, and a single bias node. Stochastic gradient descent (including momentum terms) is used to minimize the cross-entropy loss on mini-batches of size $64$ selected at random. The initial learning rate is set to $0.01$ and then multiplied by $0.99$ every $5$ epochs.
For each set, we train three random initializations. 
Note that we train the \textbf{MNISTlc} models for $1\ 000$ epochs and models from the other two sets for $100$ epochs. This is because the label-corrupted dataset required more epochs to interpolate.

\begin{figure}[b]
    \centering
    \includegraphics[width=0.48\linewidth]{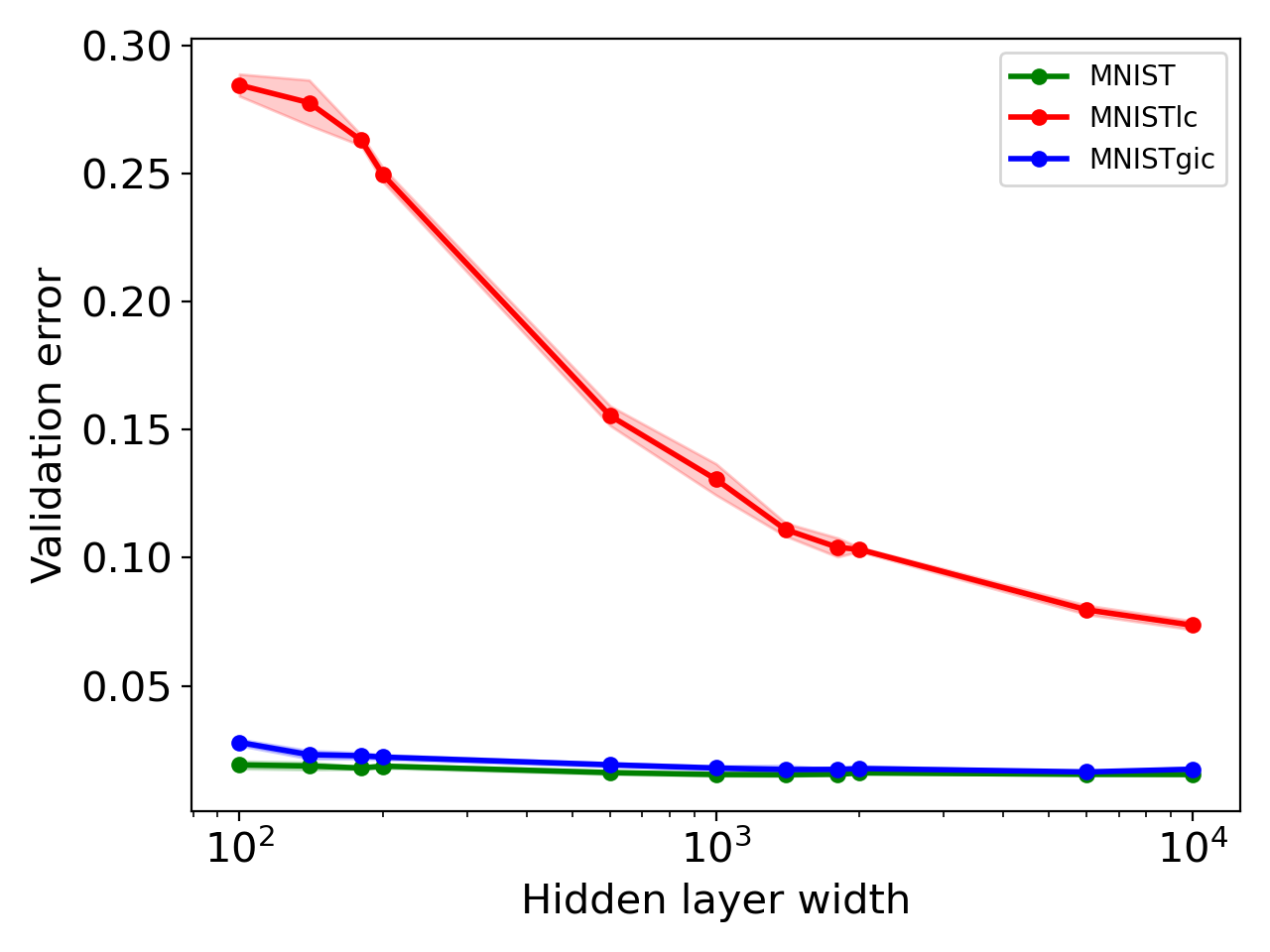}
    \includegraphics[width=0.48\linewidth]{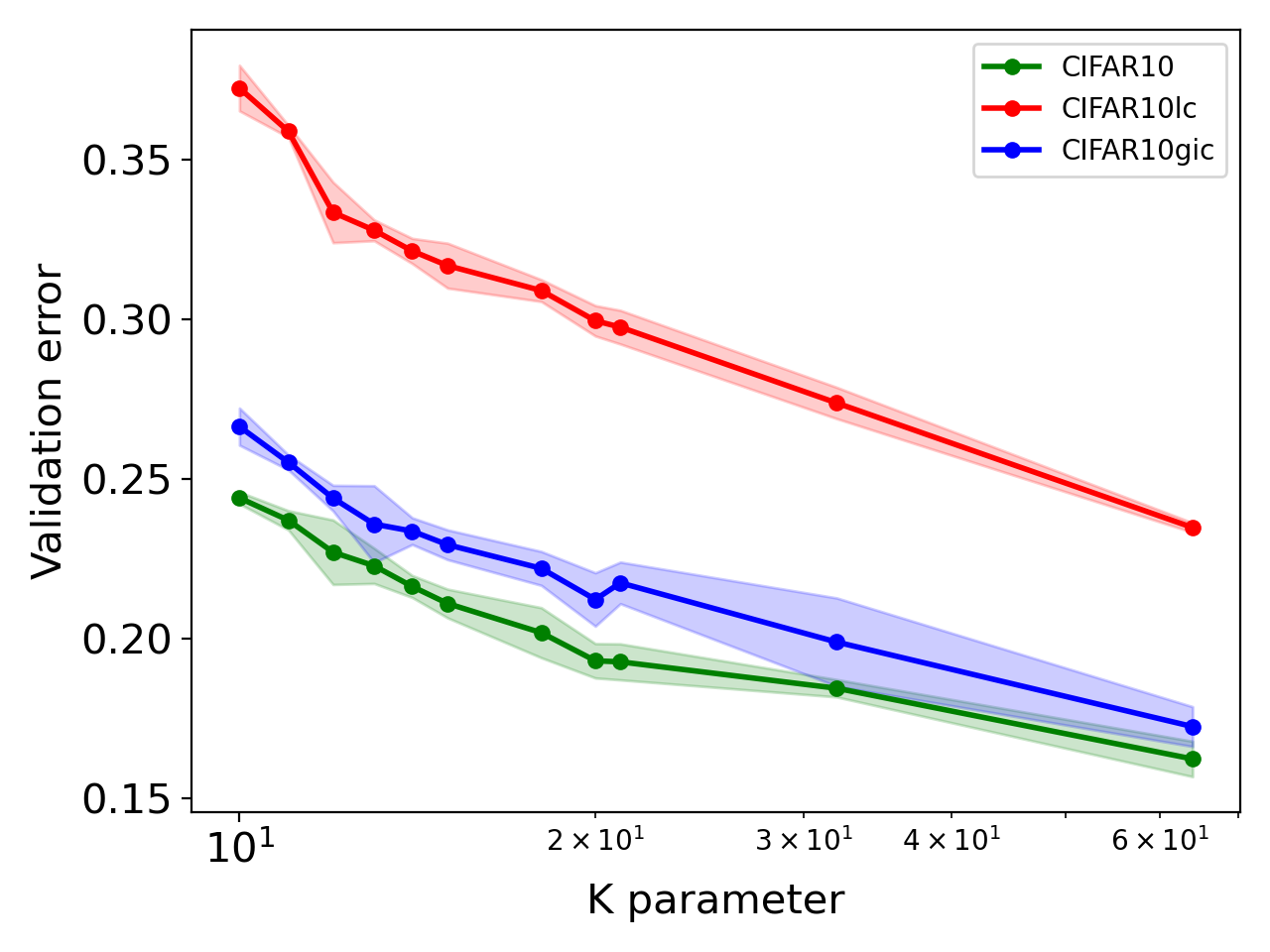}
    \caption{Validation error for MNIST models (left) and CIFAR10 models (right). Values are averaged over three random seeds and shaded areas indicate standard deviation. Note that all models interpolated except for the smallest two capacities for \textbf{MNISTlc} and \textbf{CIFAR10lc}. The maximum train error (across all models) was $0.0573$.}
    \label{fig:performance}
\end{figure}

The resulting generalization ability of all three sets are depicted in Fig.~\ref{fig:performance} (left). Note that, as expected for all three tasks, with more capacity we see an improvement in validation set performance. Also note that only the label corruption results in any significant reduction in validation performance, as also previously reported in \cite{theunissen2020benign}.

\subsection{CIFAR10 Models}

We also replicate the MNIST experiments using CNNs trained on the CIFAR10 dataset, with $10\%$ corruption (where applicable).  We use a similar architecture to the `standard' CNN used by Nakkiran et al.~\cite{nakkiran_dd}. Each CNN consists of four ReLU-activated convolutional layers, with $[k, 2k, 4k, 8k]$ output channels respectively, where we choose various values of $k$ between $10$ and $64$ to create a group of models with varying capacity.  Each model also includes max and average pooling layers, and a final fully connected layer of $400$ nodes. The three sets of CIFAR10 models \textbf{CIFAR10}, \textbf{CIFAR10lc}, and \textbf{CIFAR10gic}, refer to clean, label-corrupted, and input-corrupted models, respectively. 

All models are trained on a $45\ 000/5\ 000$ train-validation split of the CIFAR10 dataset. These models are trained for $500$ epochs in order to minimize a cross-entropy loss function on mini-batches of $256$ samples using the Adam optimizer. The initial learning rate of $0.001$ is multiplied by $0.99$ every $10$ epochs. Three initialization seeds are used. From the relevant validation errors in Fig.~\ref{fig:performance} (right), we again observe that more capacity is accompanied by better generalization performance.

\subsection{Terminology}

When describing margin behaviour, we refer to different subselections of margins based on the type of sample (clean or corrupted) as well as the type of model.
To prevent confusion, 
we always refer to these subselections using a name constructed from the type of sample and then the type of model, separated by a colon, as shown in Table.~\ref{tab:terminology}. 
For example, if we are referring to the margins for the uncorrupted samples with regard to a label-corrupted model we will refer to them as the \textit{clean:label-corrupted} margins. 

\begin{table}[h]
\centering
\caption{Sample corruption terminology.}
\label{tab:terminology}
\begin{tabular}{c|c|c|c|}
\cline{2-4}
\textit{}                                                                                          & \textbf{clean samples}         & \textbf{corrupt samples}         & \textbf{overall samples}         \\ \hline
\multicolumn{1}{|c|}{\textbf{clean model}}                                                         & \textit{clean:clean}           & N/a                              & \textit{overall:clean}                              \\ \hline
\multicolumn{1}{|c|}{\textbf{\begin{tabular}[c]{@{}c@{}}label- \\ corrupted \\ model\end{tabular}}} & \textit{clean:label-corrupted} & \textit{corrupt:label-corrupted} & \textit{overall:label-corrupted} \\ \hline
\multicolumn{1}{|c|}{\textbf{\begin{tabular}[c]{@{}c@{}}input- \\ corrupted \\ model\end{tabular}}} & \textit{clean:input-corrupted} & \textit{corrupt:input-corrupted} & \textit{overall:input-corrupted} \\ \hline
\end{tabular}
\end{table}

\section{Results}
\label{local inconsistencies}

We calculate the margins for $10\ 000$ randomly selected training samples, for all of the models defined in Section \ref{experimental setup}, using the method described in Section \ref{sec:measuring_classification_margins}. Only correctly classified samples are considered. 
This amounts to solving $11 \text{ capacities } \times 3 \text{ random seeds } \times 3 \text{ datasets } \times 9 \text{ class pairs } \times 10k \text{ samples } = 8\ 910k$ individual CMPs for both the MNIST and CIFAR10 models.  In order to solve such a large number of CMPs we utilize $240$ CPU cores split over $10$ servers, by making use of GNU-Parallel~\cite{gnu_parallel}. Next, we investigate the central tendencies of the clean and corrupt samples separately, as capacity increases (Section \ref{local}) and discuss possible implications of our observations (Section \ref{implications}).

\subsection{Local Inconsistencies}
\label{local}

\begin{figure}[t]
    \centering
    \includegraphics[width=0.49\linewidth]{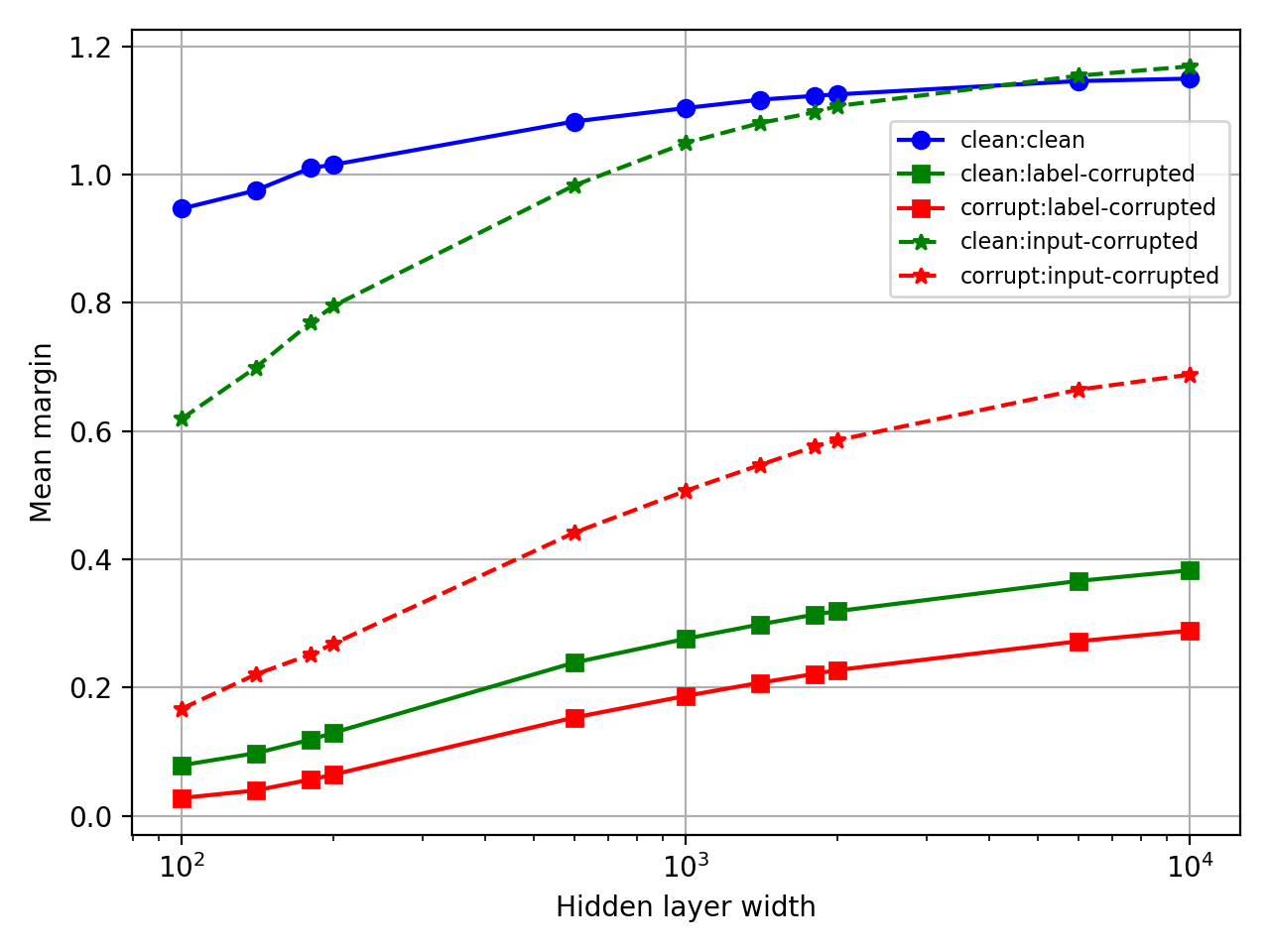}
    \includegraphics[width=0.49\linewidth]{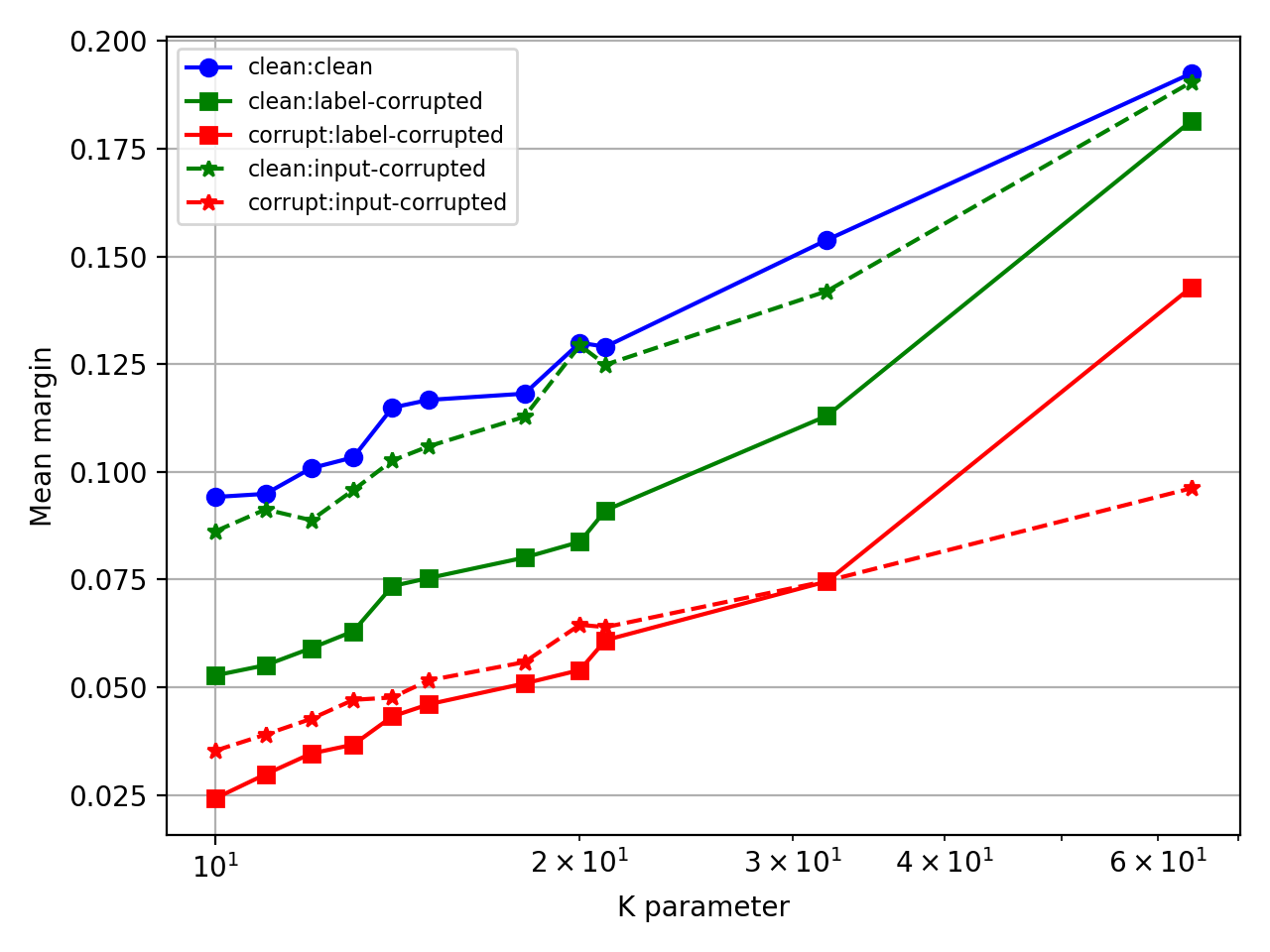}
    \caption{Mean margins for MNIST models (left) and CIFAR10 models (right).}
    \label{fig:margin_means_all_mlp_models}
\end{figure}

Fig.~\ref{fig:margin_means_all_mlp_models} shows the mean margins as a function of model capacity. We note that expected margins tend to increase along with capacity, in all cases for all types of samples. This is consistent with what we expect when using margins as an indicator of generalization.

However, we see that noise-corrupted and clean data demonstrate different tendencies: 
we see that margins for both \textit{corrupt:label-corrupted} and \textit{corrupt:input-corrupted} samples tend to be significantly smaller than for the clean samples in the same models at the same capacity.
Furthermore, we observe that the \textit{clean:clean} and \textit{clean:input-corrupted} margins are similar, especially at higher capacities, while \textit{clean:label-corrupted} margins are seen to be significantly smaller.

It is interesting that the average margins, measured on CIFAR10, are smaller than the MNIST ones. The CIFAR10 features contain $3\ 072$ dimensions and MNIST only $784$. We expect Euclidean distance in higher dimensions to be larger. We suspect that the reason for this contradiction is a large degree of inherent inter-class overlap in CIFAR10 that results in the observed small margins.

Another discrepancy between the MNIST and CIFAR10 results is that the \textit{corrupt:input-corrupted} margins for CIFAR10 models are comparable to the \textit{corrupt:label-corrupted} margins. For MNIST the \textit{corrupt:input-corrupted} margins are significantly larger than the  \textit{corrupt:label-corrupted} margins. This observation is important to note when we take a deeper look at the underlying mechanisms (in Section~\ref{deeper look}).

\begin{figure}[!ht]
    \centering
    \includegraphics[width=0.49\linewidth]{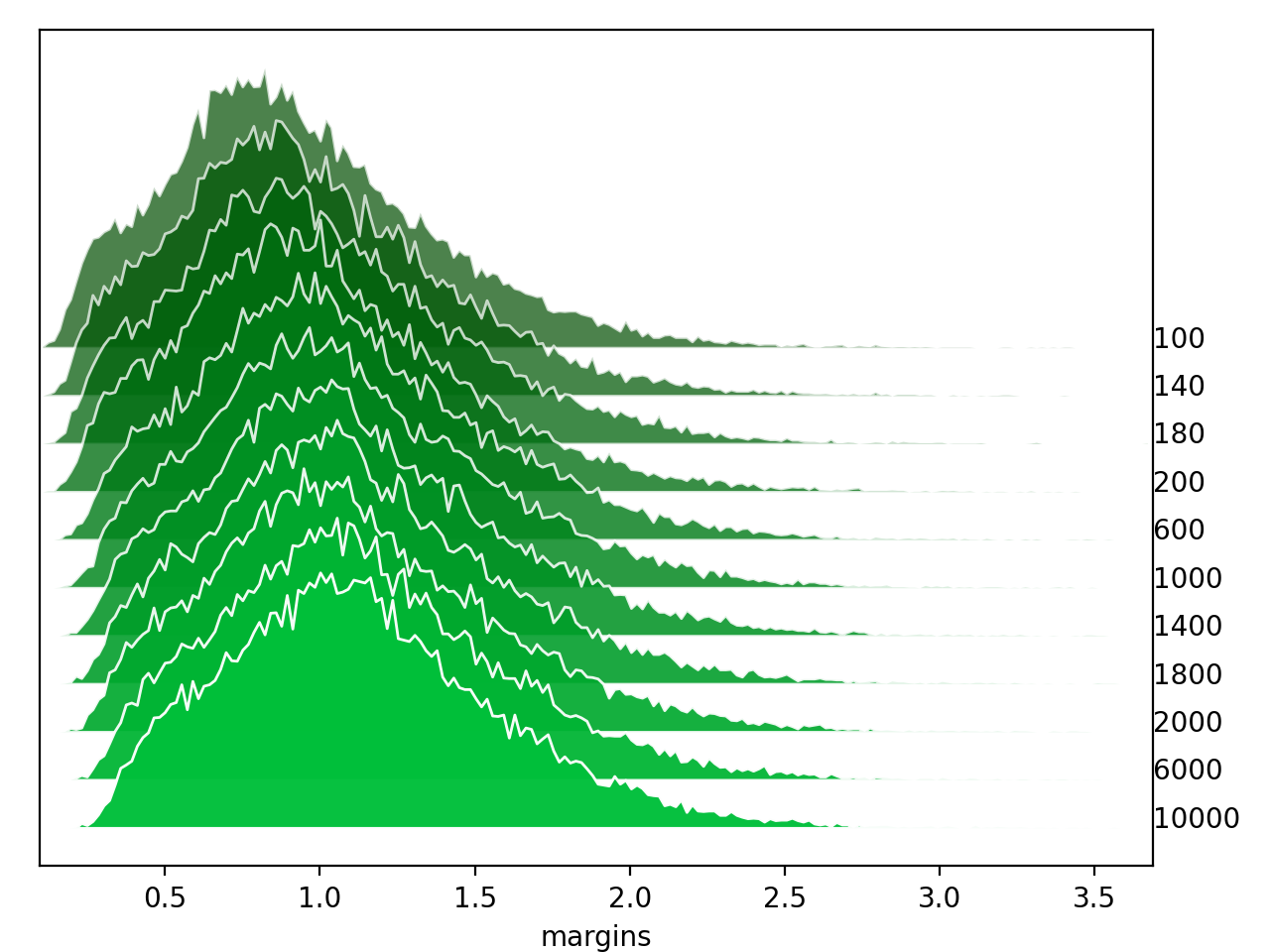}
    \includegraphics[width=0.49\linewidth]{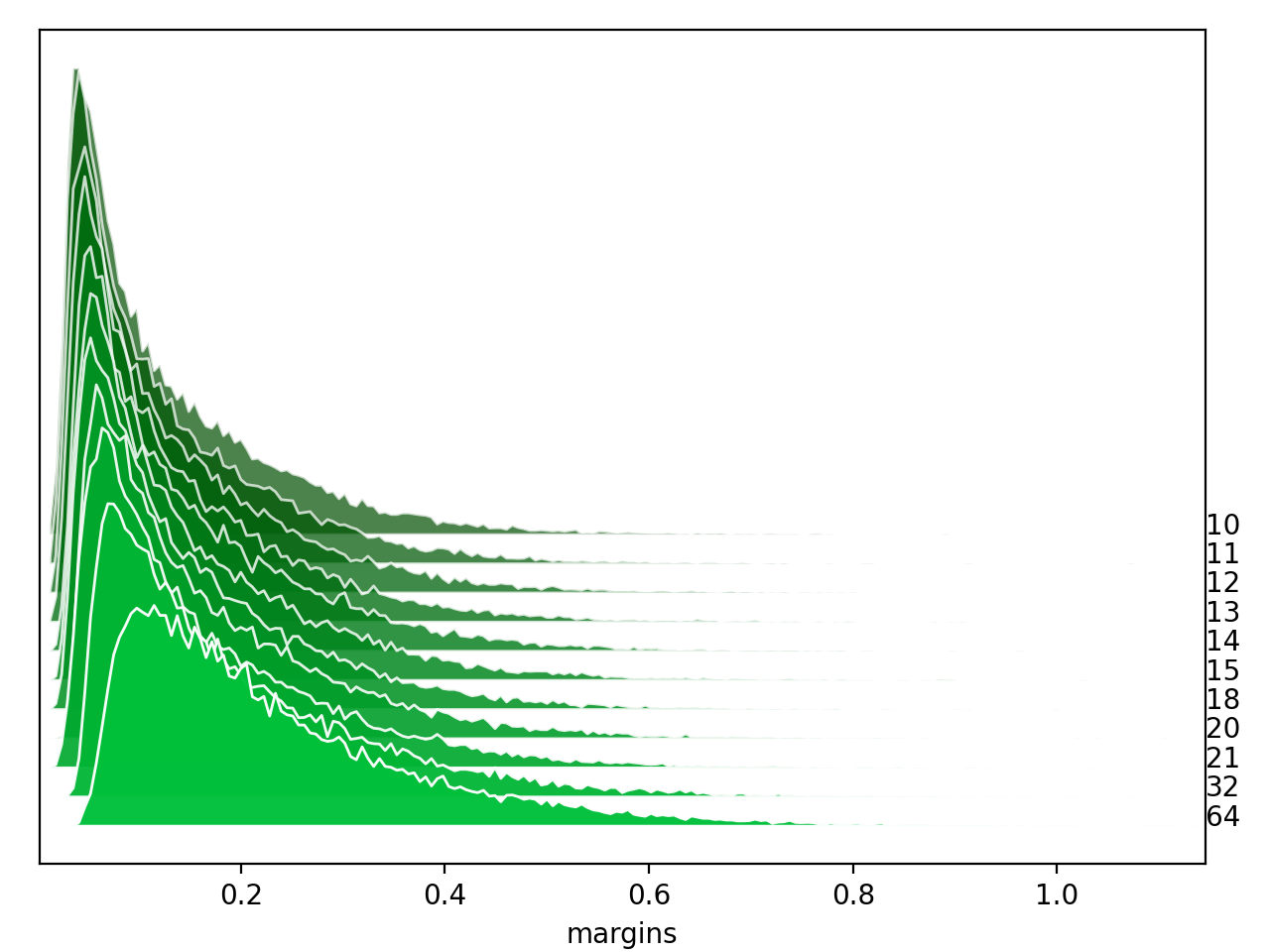}
    \includegraphics[width=0.49\linewidth]{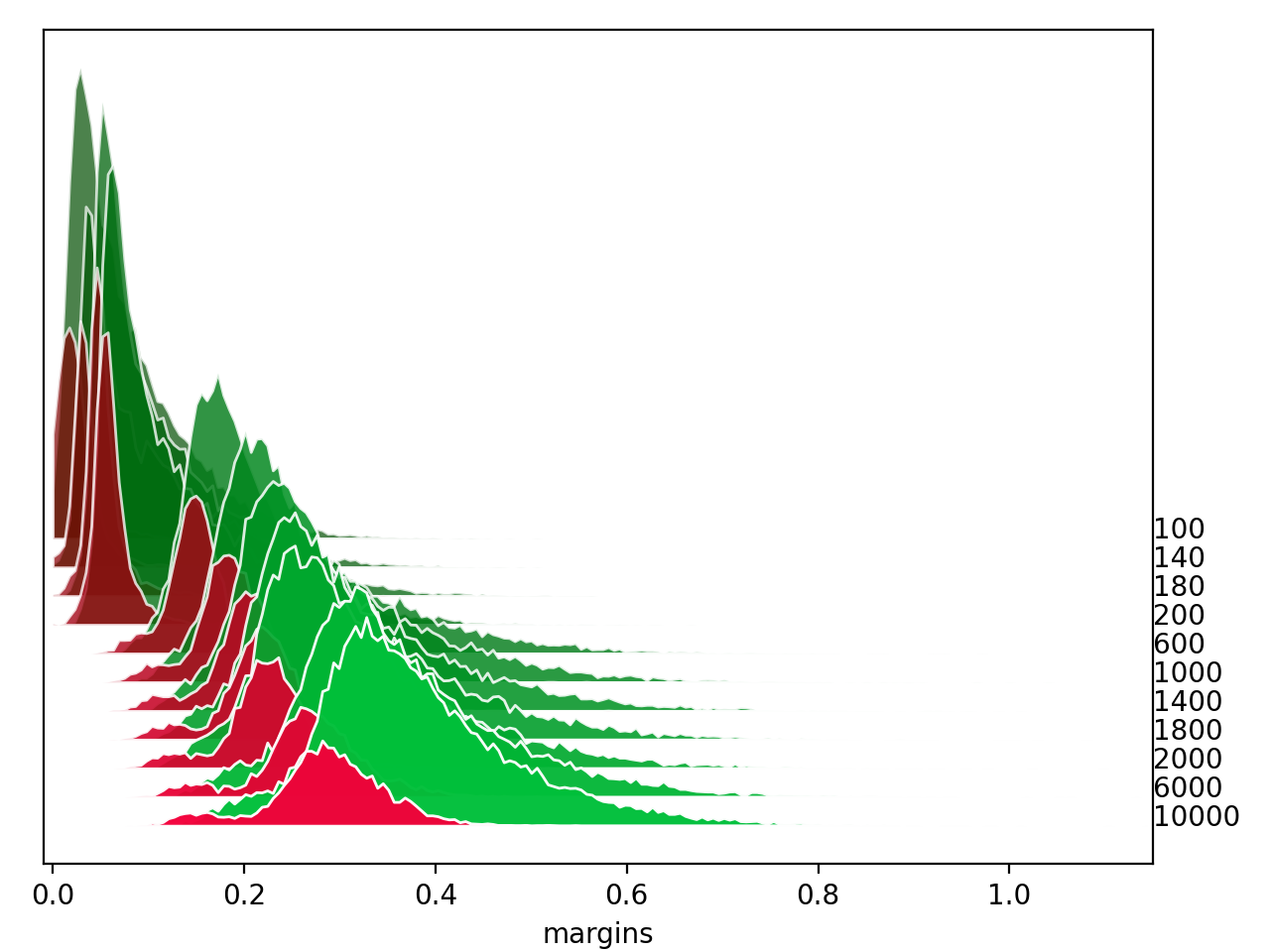}
    \includegraphics[width=0.49\linewidth]{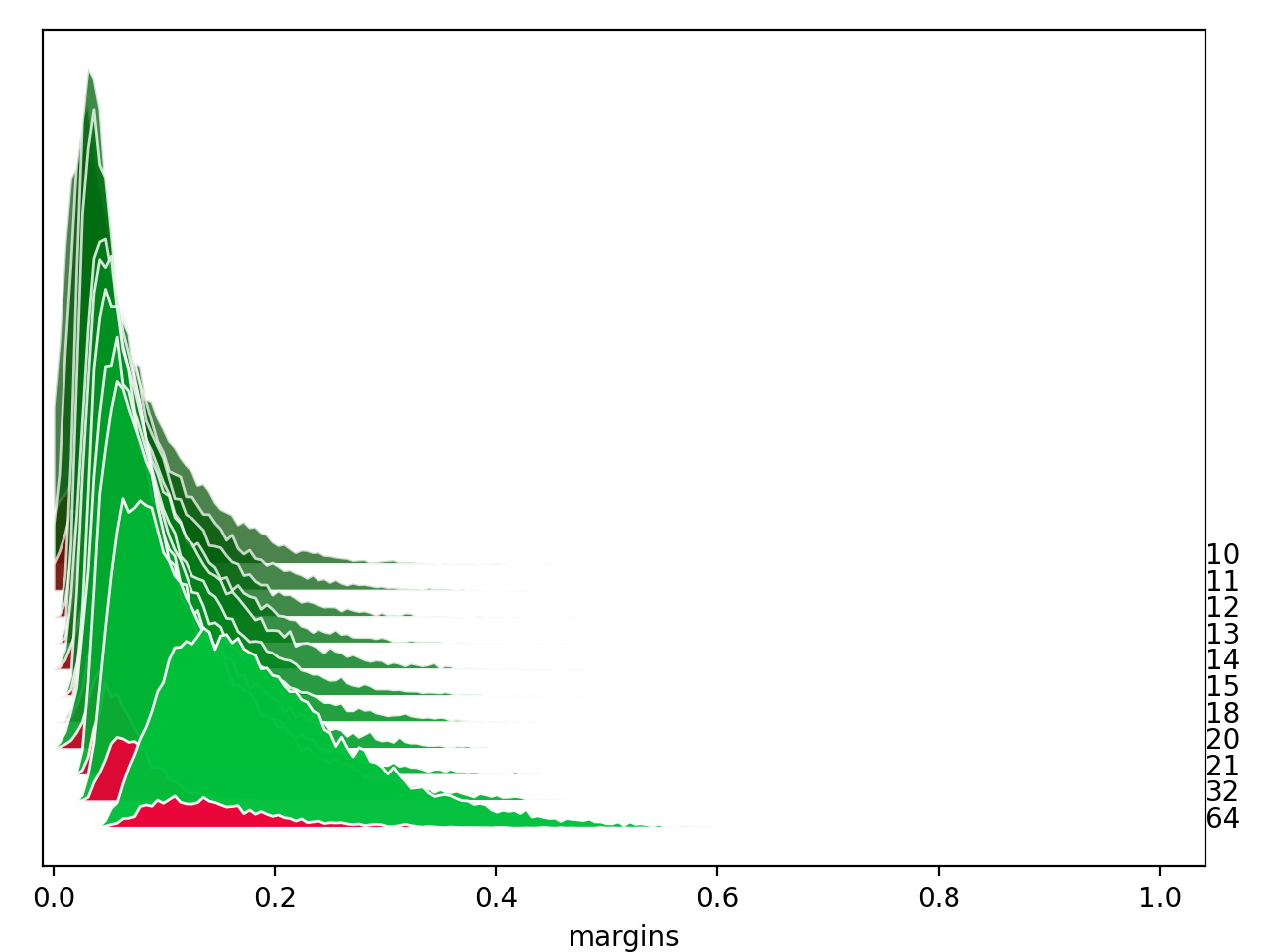}
    \includegraphics[width=0.49\linewidth]{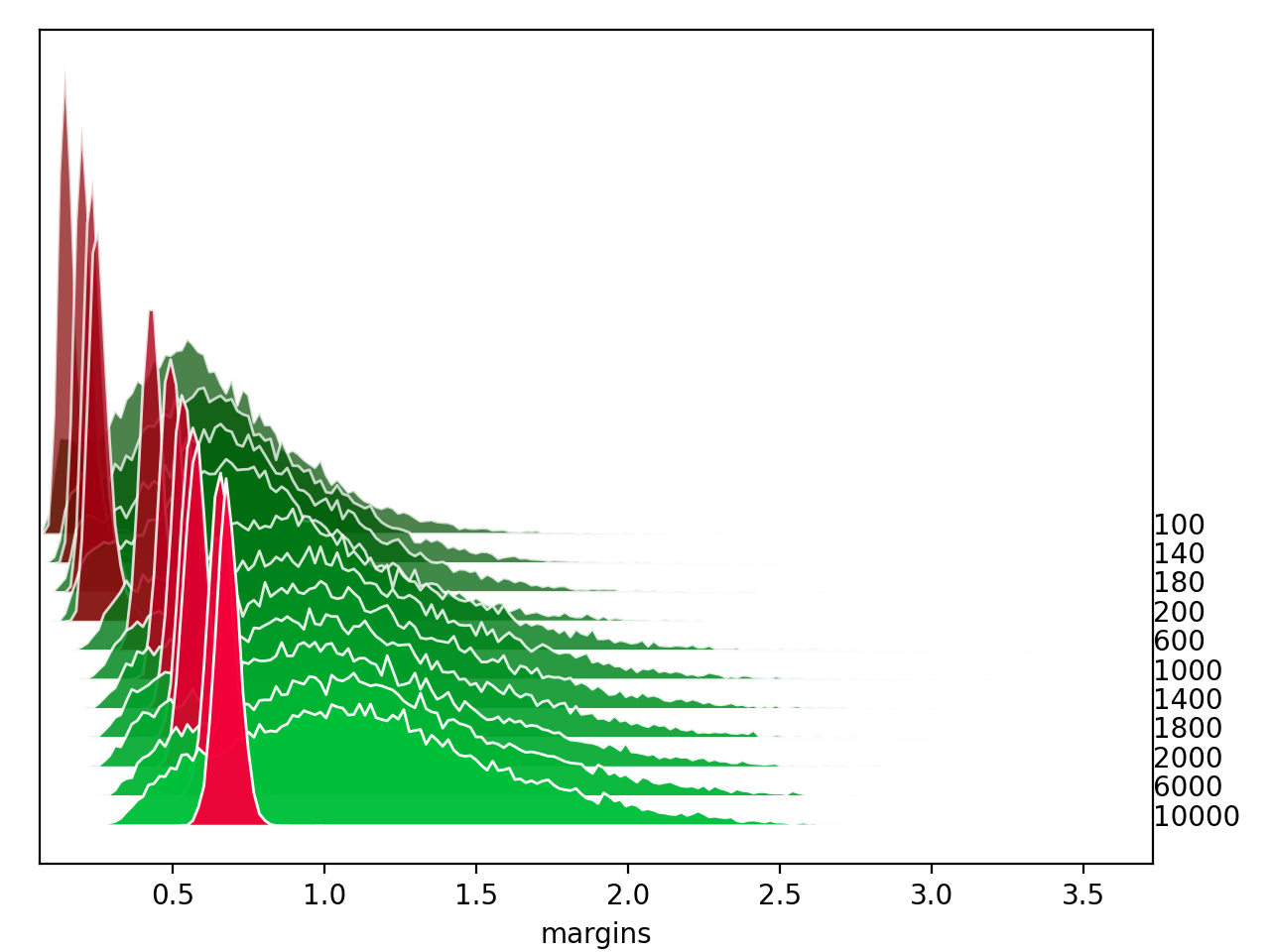}
    \includegraphics[width=0.49\linewidth]{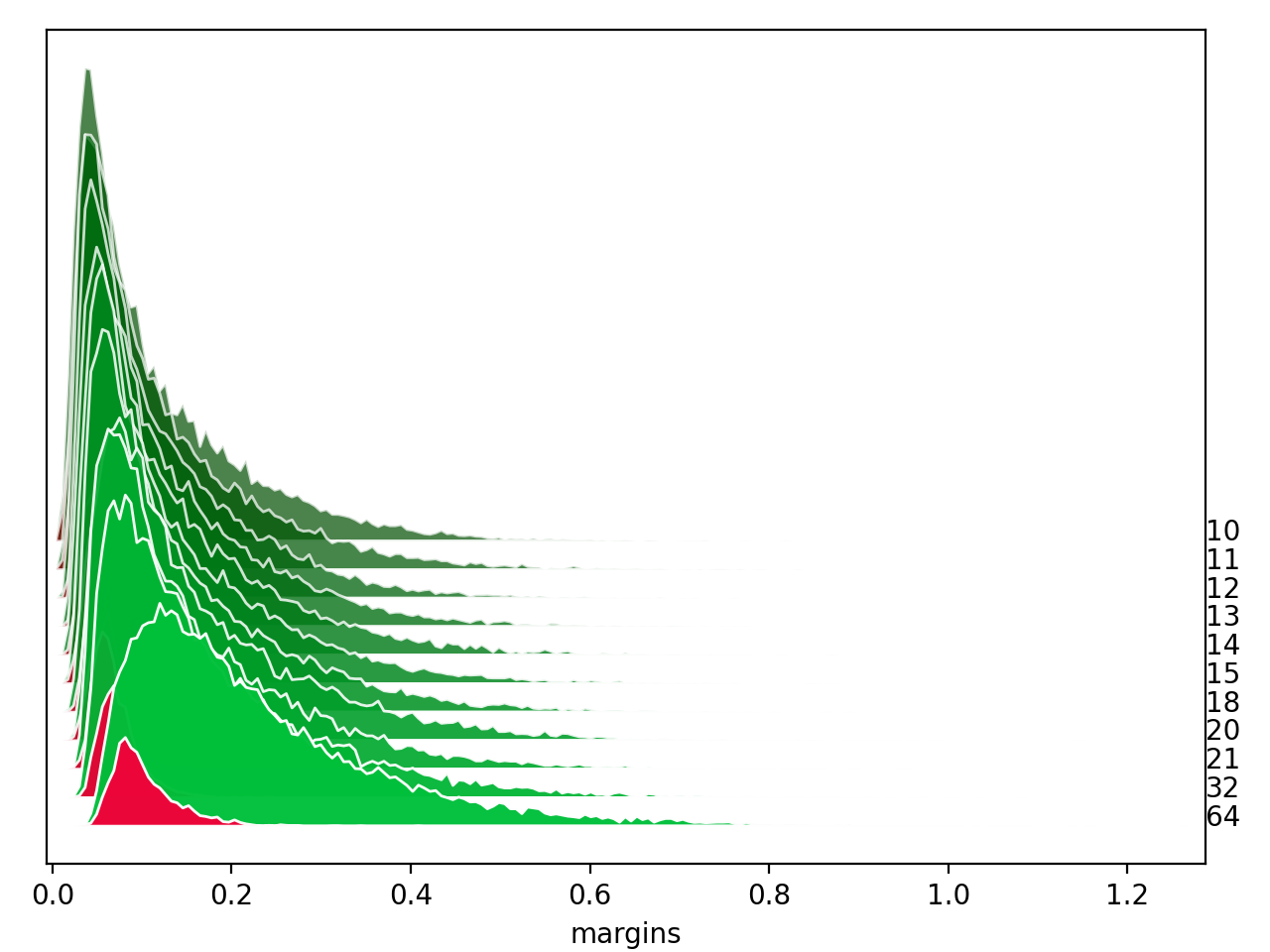}
    \caption{Margin distributions for MNIST models (left) and CIFAR10 models (right) trained on clean (top), label-corrupted (middle) and input-corrupted (bottom) train sets. Within each plot, from top to bottom, distributions are ordered by ascending model size. The relevant capacity metric is shown on the right. Green and red distributions are constructed from clean and corrupted samples, respectively.}
    \label{fig:margin distributions}
\end{figure}

Next, we look at the distributions of margins underlying the means in Fig.~\ref{fig:margin_means_all_mlp_models}. These results are shown in Fig.~\ref{fig:margin distributions}. We construct histograms of the margins measured at each capacity. These histograms share a common set of bins on the horizontal axes.

We see that most of these distributions are right-skewed distributions with a long tail containing relatively large margins. This indicates that the mean might be a slightly overestimated measure of the central tendency of margins. The margins for the \textbf{CIFAR10lc} (right-middle) set show similar trends to that of the \textbf{MNISTlc} set (left-middle), but the distributions are even more right-skewed, indicating that their central tendencies are even smaller than the means in Fig.~\ref{fig:margin_means_all_mlp_models} (right) suggest.

We also see that the \textit{corrupt:input-corrupted} margin distributions of the \textbf{MNISTgic} set (left-bottom) are normally distributed with a relatively low variance, compared to the other distributions. This suggests that all models are constructing similar decision boundaries around \textit{corrupt:input-corrupted} samples. There is not much diversity in how far samples tend to be from their nearest boundary.
The \textit{corrupt:label-corrupted} margins for the \textbf{MNISTlc} set (left-middle), on the other hand, show much higher diversity. The shape of the distribution changes drastically as capacity increases. At the critically small capacities we see a distribution resembling the \textit{corrupt:input-corrupted} margin distributions and at higher capacities some \textit{corrupt:label-corrupted} samples obtain almost outlying small margins.

\subsection{Discussion}
\label{implications}

We notice a few key inconsistencies regarding the use of a global average margin as an indicator of the extent to which a model separates samples by means of a decision boundary:
\begin{enumerate}
    \item Samples that are on-manifold, but problematic in terms of their class separability (of which label-corrupted samples are extreme examples) have much lower margins than the global mean would suggest.
    \item Samples that are off-manifold and remote (of which the input-corrupted samples are extreme examples) also have much smaller margins, even though these samples do not affect the generalization ability of the model significantly.
    \item In general, the margin distributions are not normally distributed. A mean might be skewed by the fact that the margin distributions have long tails containing extremely large margins values.
\end{enumerate}

Given these inconsistencies, we can ask whether the global average margin metrics, which are often used to predict or promote generalization, are sound. It seems that the corrupted margins tend to increase in proportion with the clean margins (see Fig.~\ref{fig:margin_means_all_mlp_models}). Assuming the models to be compared fitted exactly the same training samples, this implies that an average margin will only work if the two models for which margins are being compared contain an approximately equal number of samples with these distinct locally inconsistent margin behaviours. If a small set of samples are averaged over, this could become a problem. If the models to be compared have varying training set performance or were not trained on exactly the same training set this could become an even more significant problem.

It can also be argued that margin-based generalization predictors are more sensitive to off-manifold noise than to on-manifold noise. That is because the small \textit{corrupt:label-corrupted} margins rightly indicate the poor generalization of label-corrupted models. However, the small \textit{corrupt:input-corrupted} margins erroneously also indicate poor generalization.

\section{A Deeper Look}
\label{deeper look}

Now that we have discussed the observed inconsistencies and their possible implications, we explore the origin of these inconsistencies further. We do this by posing three concrete questions based on the results in the previous section. 
After each question, we propose possible answers, producing additional measurements where these can shed light on the underlying phenomena.

\textbf{(1) Why are the \textit{overall:label-corrupted} margins so small?} We note that label corruption is expected to result in many samples that have different targets while being close to each other in the input space. One can think of a sample's minimum distance to another sample with a different target as its absolute maximum possible margin since a boundary needs to be drawn between them, assuming both have been correctly classified during training. We propose that this is the main factor contributing to the small \textit{overall:label-corrupted} margins.

\begin{figure}[h]
    \centering
    \includegraphics[width=0.49\linewidth]{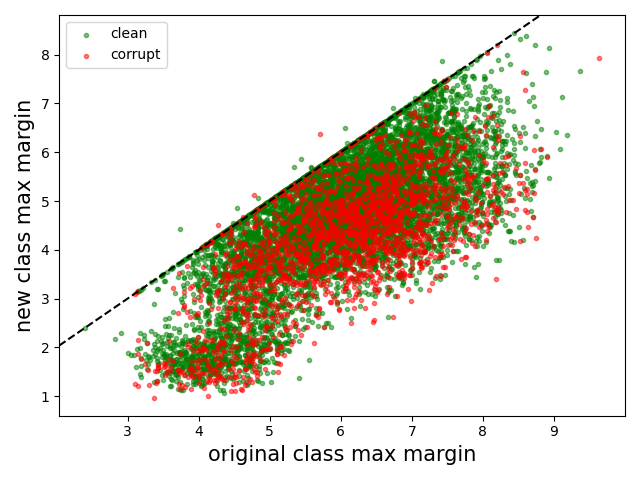}
    \includegraphics[width=0.49\linewidth]{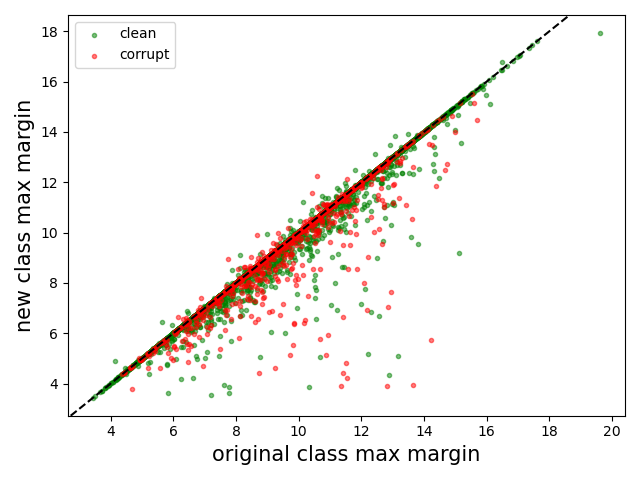}
    \caption{Maximum margins before vs. after label corruption for \textbf{MNISTlc} (left) and \textbf{CIFAR10lc} (right). Green points represent clean samples and red points represent corrupt samples. The dashed line indicates $y = x$.}
    \label{fig:max margins lc}
\end{figure}

To test this hypothesis, we randomly select $10\ 000$ training samples from the data that the models in the \textbf{MNISTlc} and \textbf{CIFAR10lc} sets are trained on. We then measure the `maximum margin' as the minimum Euclidean distance between each sample $(\mathbf{x}_{1}, c_{1})$ and its nearest neighbour $(\mathbf{x}_{2}, c_{2})$ (selected from the entire train set) so that:
\begin{equation}
    \min_{\mathbf{x}_{2}} |\mathbf{x}_{1} - \mathbf{x}_{2}|_{2},~~ c_{1} \neq c_{2}
\end{equation}
We do this before and after the data is label corrupted. We then construct a scatter plot of these $10\ 000$ training samples with the distance as measured with the original targets on the horizontal axis and the potentially corrupted targets on the vertical axis. The resulting scatter plots are shown in Fig.~\ref{fig:max margins lc}.
All samples below the provided identity function line had their max margin reduced due to label corruption. Note that, as expected, the presence of label corruption causes many samples, corrupted and clean, to have drastically reduced upper bounds to their margins.

\begin{figure}[t]
    \centering
    \includegraphics[width=0.8\linewidth]{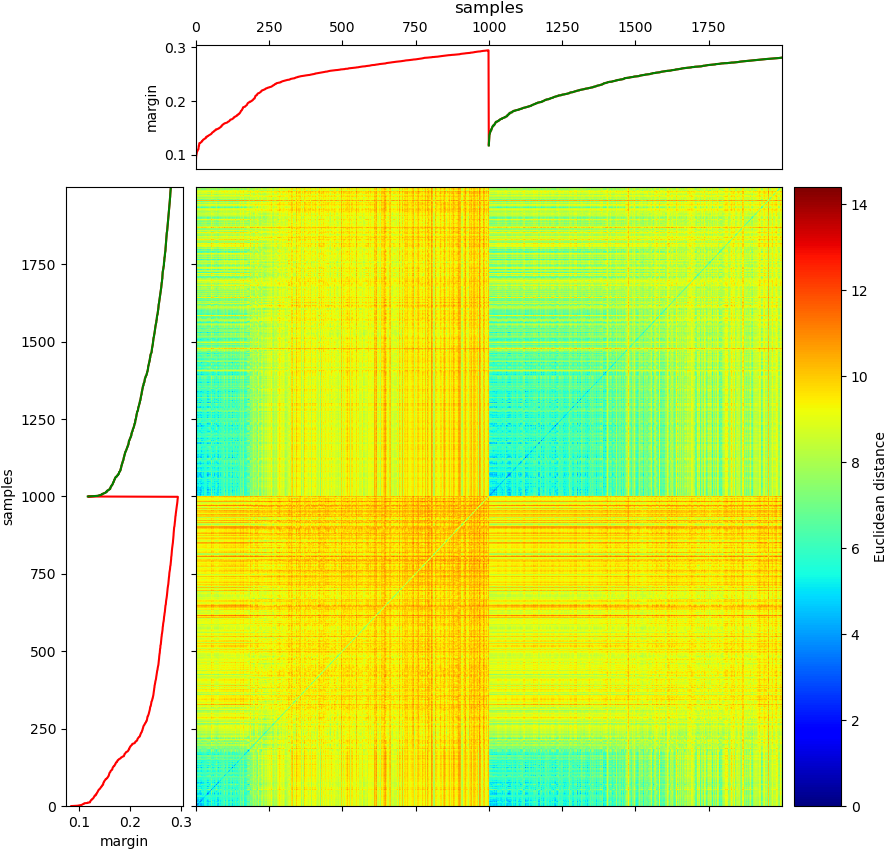}
    \caption{Euclidean distance between $1\ 000$ \textit{clean:label-corrupted} and $1\ 000$ \textit{corrupt:label-corrupted} samples with the smallest margins for a $1 \times 10\ 000$ model \textbf{MNISTlc}. Entries in the dissimilarity matrix are ordered by corruption status first, then by margin size. See the corresponding margins in the axis plots for the exact margins. Red curves refer to corrupted margins and green curves refer to clean margins.}
    \label{fig:proximity lc}
\end{figure}

If our hypothesis is true then the smallest margins should correspond to the \textit{clean:label-corrupted} and \textit{corrupt:label-corrupted} samples that are the closest to each other in the input space. We confirm this by constructing a Euclidean distance matrix comparing the $1\ 000$ \textit{clean:label-corrupted} samples with the smallest margins and the $1\ 000$ \textit{corrupt:label-corrupted} samples with the smallest margins for one of the biggest models from the \textbf{MNISTlc} set. This is presented in Fig.~\ref{fig:proximity lc}.

Note that it is the samples that are relatively close to each other that tend to have the smallest margins. From these results we conclude that a significant factor leading to the small margins observed in label-corrupted models, is the proximity (in the input space) of samples with different targets. This also accounts for the observation that \textit{corrupt:label-corrupted} margins tend to be smaller than \textit{clean:label-corrupted} margins. There are more \textit{clean:label-corrupted} samples than \textit{corrupt:label-corrupted} samples. Therefore, fewer clean samples are moved closer to a different target sample. The result is that fewer \textit{clean:label-corrupted} margins are reduced.

\begin{figure}[t]
    
    \centering
    \includegraphics[width=0.8\linewidth]{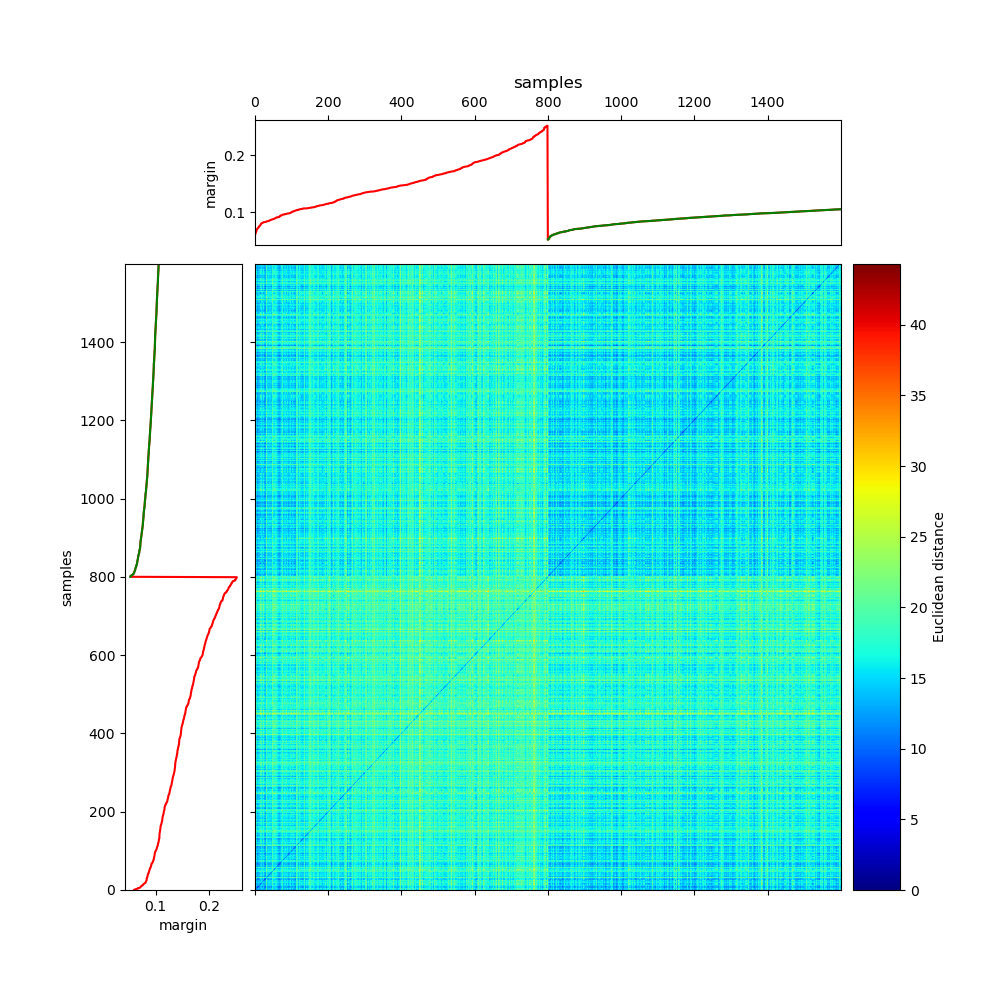}
    \caption{Euclidean distance between $1\ 000$ \textit{clean:label-corrupted} and $1\ 000$ \textit{corrupt:label-corrupted} samples with the smallest margins for a $k = 64$ model \textbf{CIFAR10lc}. Entries in the dissimilarity matrix are ordered by corruption status first, then by margin size. See the corresponding margins in the axis plots for the exact margins. Red curves refer to corrupted margins and green curves refer to clean margins.}
    \label{fig:proximity lc cifar10}
    
\end{figure}

We performed the same analysis on one of the largest \textbf{CIFAR10lc} models in Fig.~\ref{fig:proximity lc cifar10}. Note that the phenomenon is less clear here, but this is to be expected when considering the relatively few samples that have had their maximum margins reduced significantly in the CIFAR10 case, see Fig.~\ref{fig:max margins lc} (right).

\begin{figure}[t]
    \centering
    \includegraphics[width=0.49\linewidth]{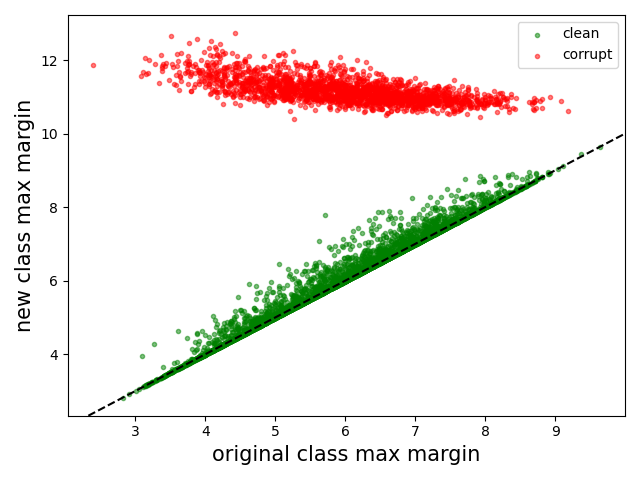}
    \includegraphics[width=0.49\linewidth]{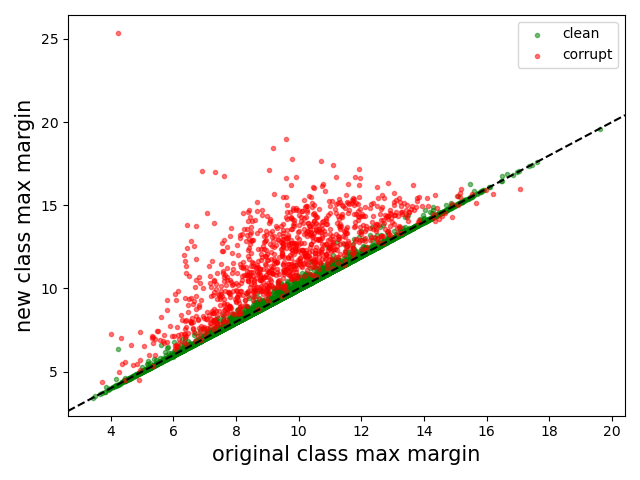}
    \caption{Maximum margins before vs. after corruption for \textbf{MNISTgic} (left) and \textbf{CIFAR10gic} (right). Green points represent clean samples and red points represent corrupt samples. The dashed line indicates $y = x$.}
    \label{fig:max margins gaussian}
\end{figure}

\textbf{(2) Why are the \textit{corrupt:input-corrupted} margins smaller than the \textit{clean:input-corrupted} margins?} To determine whether a similar phenomenon is reducing the \textit{corrupt:input-corrupted} margins we generate a similar scatter plot to Fig.~\ref{fig:max margins lc} but for the \textbf{MNISTgic} and \textbf{CIFAR10gic} sets. This is seen in Fig.~\ref{fig:max margins gaussian}.
In contrast to the \textit{overall:label-corrupted} samples we see that virtually all \textit{overall:input-corrupted} samples have either increased or unchanged maximum margins. Strikingly, the \textit{corrupt:input-corrupted} samples, for the \textbf{MNISTgic} set, have extremely high maximum margins.

This is an apparent contradiction. If the proximity of different-target samples reduces the average margin and \textit{corrupt:input-corrupted} samples are extremely distant, at least in the \textbf{MNISTgic} case, from any other sample (minimum of $10$ in Fig.~\ref{fig:max margins gaussian}), why are \textit{corrupt:input-corrupted} margins small? And why are \textit{clean:input-corrupted} margins slightly smaller than \textit{clean:clean} margins, when they have slightly larger maximum margins?

Regrettably, we do not have as strong an argument for the inconsistency 
regarding input-corrupted samples as we do for label corruption. We speculate that the relatively small \textit{corrupt:input-corrupted} margins are a result of the remoteness of these samples. They are so far off manifold and far from each other that there is little incentive to increase their respective margins beyond a certain model-specific maximum. The previously-mentioned lack of variance in margins we observe for \textit{corrupt:input-corrupted} margins in Fig.~\ref{fig:margin distributions} supports this notion.

With regards to the \textbf{CIFAR10gic}  maximum margins, Fig.~\ref{fig:max margins gaussian} (right), we observe that the \textit{corrupt:input-corrupted} samples do not undergo such a drastic increase in maximum margin after input corruption. Consequently, we deduce that they are not as remote (and off manifold) as their MNIST counterparts. This deduction is consistent with two observations we made earlier:
\begin{itemize}
    \item (See Fig.~\ref{fig:performance}): \textbf{CIFAR10gic} models achieve a noticeably worse generalization ability than \textbf{CIFAR10} models. This is in contrast to the effect of input corruption in the MNIST models.
    \item (See Fig.~\ref{fig:margin_means_all_mlp_models}): For the CIFAR10 models, \textit{corrupt:input-corrupted} margins are similar to \textit{corrupt:label-corrupted} margins, again in contrast to corresponding MNIST models.
\end{itemize}

\textbf{(3) Why are \textit{clean:input-corrupted} margins smaller than \textit{clean:clean} margins?} Without the \textit{corrupt:input-corrupted} samples in close proximity to the \textit{clean:input-corrupted} samples we might expect them to have similar margins to \textit{clean:clean}, across all capacities. However, we see that they tend to be slightly smaller. 
We hypothesize that this is a result of the capacity it requires to fit \textit{corrupt:input-corrupted} samples. It is known that these samples are fitted later and require more capacity than clean samples~\cite{krueger2017deep,theunissen2020benign}, and that margins tend to increase with capacity. It is then reasonable to conclude that the lack of \textit{clean:input-corrupted} margin size is a result of the lack of available capacity. This idea is supported by Fig.~\ref{fig:margin_means_all_mlp_models} (left), where we observe that the difference between average \textit{clean:clean} margins and average \textit{clean:input-corrupted} margins decreases with added capacity, and disappears completely when models become large enough. 

\textbf{What is the takeaway?} To summarize, there seems to be two main mechanisms contributing to the observed local inconsistencies in margin behaviour.
\begin{enumerate}
    \item Samples that are very close to different-target samples will inevitably have reduced margins. This kind of reduced margin is indicative of poor generalization because it is very likely to pertain to in-distribution and on-manifold regions of feature space that are difficult to model.
    \item Samples that are extremely remote, being very distant from any other sample, will also obtain reduced margins, due to a lack of incentive to increase them. This kind of reduced margin is not indicative of poor generalization because it is likely to pertain to out-of-distribution and off-manifold regions of feature space.
\end{enumerate}

\section{Conclusion}

In this work we show that some training samples are consistently modeled with small margins while affecting generalization in different ways. Specifically, we find that samples representing on-manifold corruption (e.g. label corruption) and off-manifold corruption (e.g. Gaussian input corruption) both 
have margins that are smaller than those of uncorrupted training samples,
though only the presence of the former significantly affects generalization. 

This is a novel observation of a phenomenon that will require consideration if margins are used in methods to predict generalization of ANNs. We use label and Gaussian input corruption as tools but hypothesize that similar behaviour is possible in natural datasets that contain significant inter-class overlap in the input space, or a large portion of off-manifold samples, respectively. We conclude that a global average margin will be more useful in predicting generalization if it considers these local inconsistencies, or contains an equal proportion of these kinds of samples for all models being compared.

In addition to providing a precise comparison of the way in which different types of margins change with increased capacity, we explore some of the possible reasons for the behaviour observed. Specifically, we find that the reduced margins accompanying label corruption (i.e. on-manifold corruption) are a result of the maximum margin (closest sample of a different class) being reduced by label corruption, for a large portion of the train set. 
For input corruption (i.e. off-manifold corruption), we speculate that the reduced margins come from a lack of incentive to increase these margins, since the input-corrupted samples are remote from other training samples.

These are hypotheses that we will be investigating further in future work. 
We also plan to extend this work to hidden layers, as hidden layer margins have been shown to similarly relate to generalization behaviour~\cite{rep_based_complexity_pgdl,predict_gen_margin}. Additionally, a more concrete understanding of the influence (or lack thereof) off-manifold samples have on the margins of on-manifold samples will be useful.

\subsubsection{Acknowledgements}

We thank and acknowledge the Centre for High Performance Computing (CHPC), South Africa, for providing computational resources to this research project.

\bibliographystyle{unsrt}
\bibliography{references}

\begin{thebibliography}{10}

\bibitem{theunissen2022missing}
Marthinus~Wilhelmus Theunissen, Coenraad Mouton, and Marelie~H Davel.
\newblock The missing margin: How sample corruption affects distance to the
  boundary in {ANN}s.
\newblock In {\em Artificial Intelligence Research: Third Southern African
  Conference, SACAIR 2022, Stellenbosch, South Africa, December 5--9, 2022,
  Proceedings}, pages 78--92. Springer, 2022.

\bibitem{vapnik1999overview}
Vladimir~N Vapnik.
\newblock An overview of statistical learning theory.
\newblock {\em IEEE transactions on neural networks}, 10(5):988--999, 1999.

\bibitem{hochreiter1997flat}
Sepp Hochreiter and J{\"u}rgen Schmidhuber.
\newblock Flat minima.
\newblock {\em Neural computation}, 9(1):1--42, 1997.

\bibitem{tishby2015deep}
Naftali Tishby and Noga Zaslavsky.
\newblock Deep learning and the information bottleneck principle.
\newblock In {\em 2015 ieee information theory workshop (itw)}, pages 1--5.
  IEEE, 2015.

\bibitem{bousquet2000algorithmic}
Olivier Bousquet and Andr{\'e} Elisseeff.
\newblock Algorithmic stability and generalization performance.
\newblock {\em Advances in Neural Information Processing Systems}, 13, 2000.

\bibitem{predict_gen_margin}
Yiding Jiang, Dilip Krishnan, Hossein Mobahi, and Samy Bengio.
\newblock {Predicting the Generalization Gap in Deep Networks with Margin
  Distributions}.
\newblock In {\em {International Conference on Learning Representations
  (ICLR)}}, 2018.

\bibitem{rep_based_complexity_pgdl}
Parth Natekar and Manik Sharma.
\newblock {R}epresentation based complexity measures for predicting
  generalization in deep learning.
\newblock {\em arXiv preprint arXiv:2012.02775}, 2020.

\bibitem{moosavi2016deepfool}
Seyed-Mohsen Moosavi-Dezfooli, Alhussein Fawzi, and Pascal Frossard.
\newblock Deepfool: a simple and accurate method to fool deep neural networks.
\newblock In {\em Proceedings of the IEEE conference on computer vision and
  pattern recognition}, pages 2574--2582, 2016.

\bibitem{boser1992training}
Bernhard~E Boser, Isabelle~M Guyon, and Vladimir~N Vapnik.
\newblock A training algorithm for optimal margin classifiers.
\newblock In {\em Proceedings of the fifth annual workshop on Computational
  learning theory}, pages 144--152, 1992.

\bibitem{weinberger2009distance}
Kilian~Q Weinberger and Lawrence~K Saul.
\newblock Distance metric learning for large margin nearest neighbor
  classification.
\newblock {\em Journal of Machine Learning Research (JMLR)}, 10(2), 2009.

\bibitem{lyu2019gradient}
Kaifeng Lyu and Jian Li.
\newblock {Gradient Descent Maximizes the Margin of Homogeneous Neural
  Networks}.
\newblock In {\em International Conference on Learning Representations (ICLR)},
  2019.

\bibitem{wei2018margin}
Colin Wei, Jason Lee, Qiang Liu, and Tengyu Ma.
\newblock On the margin theory of feedforward neural networks.
\newblock 2018.

\bibitem{zhang2021understanding}
Chiyuan Zhang, Samy Bengio, Moritz Hardt, Benjamin Recht, and Oriol Vinyals.
\newblock Understanding deep learning (still) requires rethinking
  generalization.
\newblock {\em Communications of the ACM}, 64(3):107--115, 2021.

\bibitem{neyshabur2017exploring}
Behnam Neyshabur, Srinadh Bhojanapalli, David McAllester, and Nati Srebro.
\newblock Exploring generalization in deep learning.
\newblock {\em Advances in Neural Information Processing Systems}, 30, 2017.

\bibitem{nakkiran_dd}
Preetum Nakkiran, Gal Kaplun, Yamini Bansal, Tristan Yang, Boaz Barak, and Ilya
  Sutskever.
\newblock {Deep double descent: Where bigger models and more data hurt}.
\newblock {\em {Journal of Statistical Mechanics: Theory and Experiment}},
  2021(12):124003, 2021.

\bibitem{theunissen2020benign}
Marthinus~Wilhelmus Theunissen, Marelie~H Davel, and Etienne Barnard.
\newblock Benign interpolation of noise in deep learning.
\newblock {\em South African Computer Journal}, 32(2):80--101, 2020.

\bibitem{sokolic2017robust}
Jure Sokoli{\'c}, Raja Giryes, Guillermo Sapiro, and Miguel~RD Rodrigues.
\newblock Robust large margin deep neural networks.
\newblock {\em IEEE Transactions on Signal Processing}, 65(16):4265--4280,
  2017.

\bibitem{sun2015large}
Shizhao Sun, Wei Chen, Liwei Wang, and Tie-Yan Liu.
\newblock Large margin deep neural networks: Theory and algorithms.
\newblock {\em arXiv preprint arXiv:1506.05232}, 2015.

\bibitem{large_margin_dnns}
Gamaleldin Elsayed, Dilip Krishnan, Hossein Mobahi, Kevin Regan, and Samy
  Bengio.
\newblock {L}arge margin deep networks for classification.
\newblock {\em {A}dvances in {N}eural {I}nformation {P}rocessing {S}ystems},
  31, 2018.

\bibitem{constrained_optim}
Roozbeh Yousefzadeh and Dianne~P. O'Leary.
\newblock {D}eep learning interpretation: {F}lip points and homotopy methods.
\newblock In Jianfeng Lu and Rachel Ward, editors, {\em {Proceedings of The
  First Mathematical and Scientific Machine Learning Conference}}, volume 107
  of {\em {Proceedings of Machine Learning Research}}, pages 1--26. {PMLR},
  20--24 Jul 2020.

\bibitem{deep_dig}
Hamid Karimi, Tyler Derr, and Jiliang Tang.
\newblock {C}haracterizing the decision boundary of deep neural networks.
\newblock {\em arXiv preprint arXiv:1912.11460}, 2019.

\bibitem{guan_linear_interpol}
Shuyue Guan and Murray Loew.
\newblock {A}nalysis of generalizability of deep neural networks based on the
  complexity of decision boundary.
\newblock In {\em {2020 19th IEEE International Conference on Machine Learning
  and Applications (ICMLA)}}, pages 101--106. IEEE, 2020.

\bibitem{twice_dd_real_ver}
Gowthami Somepalli, Liam Fowl, Arpit Bansal, Ping Yeh-Chiang, Yehuda Dar,
  Richard Baraniuk, Micah Goldblum, and Tom Goldstein.
\newblock Can neural nets learn the same model twice? investigating
  reproducibility and double descent from the decision boundary perspective.
\newblock In {\em Proceedings of the IEEE/CVF Conference on Computer Vision and
  Pattern Recognition}, pages 13699--13708, 2022.

\bibitem{Lecun98gradient-basedlearning}
Yann LeCun, Léon Bottou, Yoshua Bengio, and Patrick Haffner.
\newblock Gradient-based learning applied to document recognition.
\newblock {\em Proceedings of the IEEE}, 86(11):2278--2324, 1998.

\bibitem{cifar}
Alex Krizhevsky.
\newblock {Learning Multiple Layers of Features from Tiny Images}.
\newblock 05 2012.

\bibitem{nlopt_lagrange_2}
Ernesto~G Birgin and Jose~Mario Martinez.
\newblock {Improving ultimate convergence of an augmented Lagrangian method}.
\newblock {\em Optimization Methods and Software}, 23(2):177--195, 2008.

\bibitem{ccsaq}
Krister Svanberg.
\newblock {A} class of globally convergent optimization methods based on
  conservative convex separable approximations.
\newblock {\em SIAM journal on optimization}, 12(2):555--573, 2002.

\bibitem{nlopt}
Steven~G. Johnson.
\newblock {The NLopt nonlinear-optimization package}.

\bibitem{northcutt2021pervasive}
Curtis Northcutt, Anish Athalye, and Jonas Mueller.
\newblock Pervasive label errors in test sets destabilize machine learning
  benchmarks.
\newblock In J.~Vanschoren and S.~Yeung, editors, {\em Proceedings of the
  Neural Information Processing Systems Track on Datasets and Benchmarks},
  volume~1, 2021.

\bibitem{gnu_parallel}
O.~Tange.
\newblock {GNU Parallel - The Command-Line Power Tool}.
\newblock {\em ;login: The USENIX Magazine}, 36(1):42--47, Feb 2011.

\bibitem{krueger2017deep}
David Krueger, Nicolas Ballas, Stanislaw Jastrzebski, Devansh Arpit, Maxinder~S
  Kanwal, Tegan Maharaj, Emmanuel Bengio, Asja Fischer, and Aaron Courville.
\newblock Deep nets don't learn via memorization.
\newblock 2017.

\end{thebibliography}

\end{document}